\documentclass{article}
\usepackage[table,dvipsnames]{xcolor}

\usepackage{microtype}
\usepackage{graphicx}
\usepackage{booktabs} 
\usepackage[most]{tcolorbox}
\usepackage[capitalize,noabbrev]{cleveref}

\crefformat{section}{#2\S#1#3}
\crefname{figure}{Fig.}{Figs.}
\crefname{table}{Tab.}{Tabs.}

\usepackage[accepted]{icml2025}

\usepackage{amsmath}
\usepackage{amssymb}
\usepackage{mathtools}
\usepackage{amsthm}
\usepackage{subcaption} 
\usepackage{comment}

\theoremstyle{definition}

\theoremstyle{remark}

\definecolor{californiagolden}{RGB}{255, 203, 5}
\definecolor{berkeleyblue}{HTML}{002676}

\usepackage[textsize=tiny]{todonotes}

\icmltitlerunning{Submission and Formatting Instructions for ICML 2025}

\begin{document}

\twocolumn[
\icmltitle{\centering
{LLMs Can Easily Learn to Reason from Demonstrations}\\ 
Structure, not content, is what matters!
}

\icmlsetsymbol{equal}{*}

\begin{icmlauthorlist}
\icmlauthor{Dacheng Li}{equal,yyy}
\icmlauthor{Shiyi Cao}{equal,yyy}
\icmlauthor{Tyler Griggs}{equal,yyy}
\icmlauthor{Shu Liu}{equal,yyy}
\icmlauthor{Xiangxi Mo}{yyy}
\icmlauthor{Eric Tang}{xxx}
\icmlauthor{Sumanth Hegde}{xxx}
\icmlauthor{Kourosh Hakhamaneshi}{xxx}
\icmlauthor{Shishir G. Patil}{yyy}
\icmlauthor{Matei Zaharia}{yyy}
\icmlauthor{Joseph E. Gonzalez}{yyy}
\icmlauthor{Ion Stoica}{yyy}
\end{icmlauthorlist}

\icmlaffiliation{xxx}{Anyscale}
\icmlaffiliation{yyy}{Department of Electrical Engineering and Computer Sciences, University of California, Berkeley}

\icmlcorrespondingauthor{Ion Stoica}{istoica@berkeley.edu}
\vskip 0.3in
]
\printAffiliationsAndNotice{\icmlEqualContribution} 




\begin{abstract}
Large reasoning models (LRMs) tackle complex reasoning problems by following long chain-of-thoughts (Long CoT) that incorporate reflection, backtracking, and self-validation. 
However, the training techniques and data requirements to elicit Long CoT remain poorly understood. 
In this work, we find that a Large Language model (LLM) can effectively learn Long CoT reasoning through data-efficient supervised fine-tuning (SFT) and parameter-efficient low-rank adaptation (LoRA). 
With just \textbf{17k} long CoT training samples, the Qwen2.5-32B-Instruct model achieves significant improvements on a wide range of math and coding benchmarks, including \textbf{56.7\%} (+40.0\%) on AIME 2024 and \textbf{57.0\%} (+8.1\%) on LiveCodeBench, competitive to the proprietary o1-preview model's score of 44.6\% and 59.1\%.

More importantly, we find that \textbf{the structure of Long CoT is critical to the learning process, whereas the content of individual reasoning steps has minimal impact}. Perturbations affecting content, such as training on incorrect samples or removing reasoning keywords, have little impact on performance. In contrast, structural modifications that disrupt logical consistency in the Long CoT, such as shuffling or deleting reasoning steps, significantly degrade accuracy. 
For example, a model trained on Long CoT samples with incorrect answers still achieves only \textbf{3.2\%} lower accuracy compared to training with fully correct samples. 
These insights deepen our understanding of how to elicit reasoning capabilities in LLMs and highlight key considerations for efficiently training the next generation of reasoning models. \color{berkeleyblue}{This is the academic paper of our previous released Sky-T1-32B-Preview model}. Codes are available at \url{https://github.com/NovaSky-AI/SkyThought}.
\end{abstract}

\section{Introduction}
\label{sec:intro}

\begin{figure*}[ht!]
    \centering
    \begin{subfigure}[b]{\textwidth}  
        \centering
        \includegraphics[width=.99\linewidth]{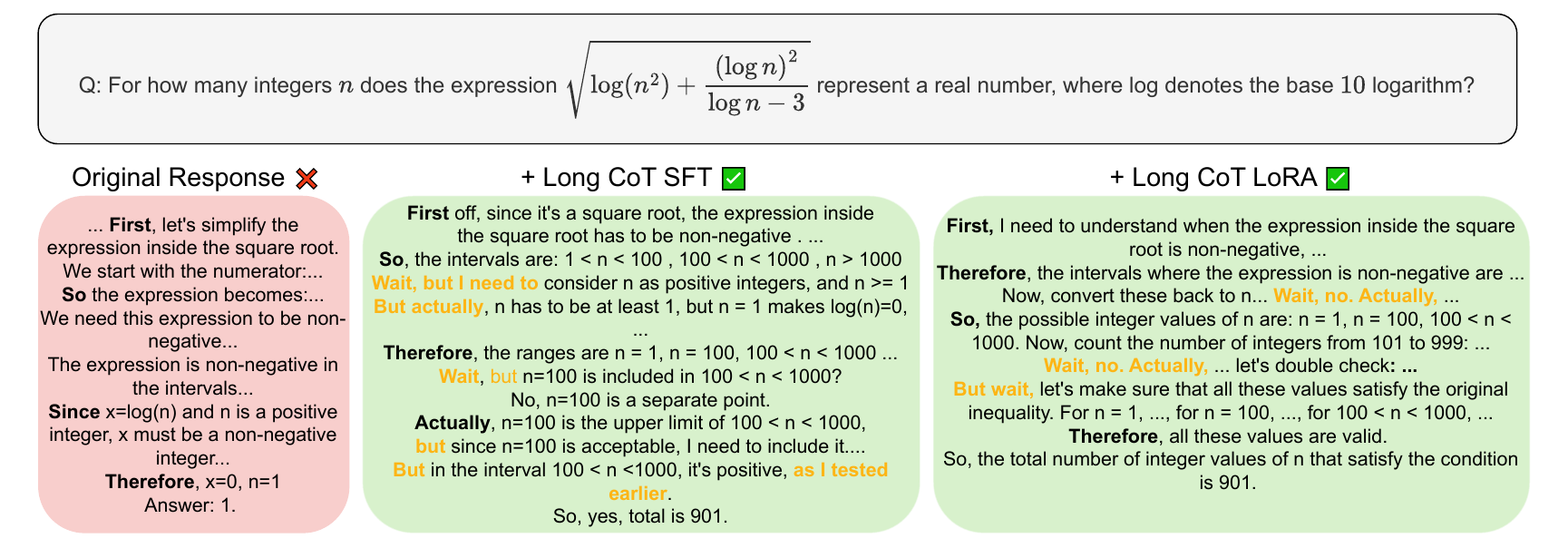}
        \caption{Responses of the base model, with Long CoT SFT, and with Long CoT LoRA.} 
        \label{fig:response_demo}
    \end{subfigure}
    \hfill
    \vspace{0.1em}
    \begin{subfigure}[b]{\textwidth}  
        \centering
        \includegraphics[width=.95\linewidth]
        {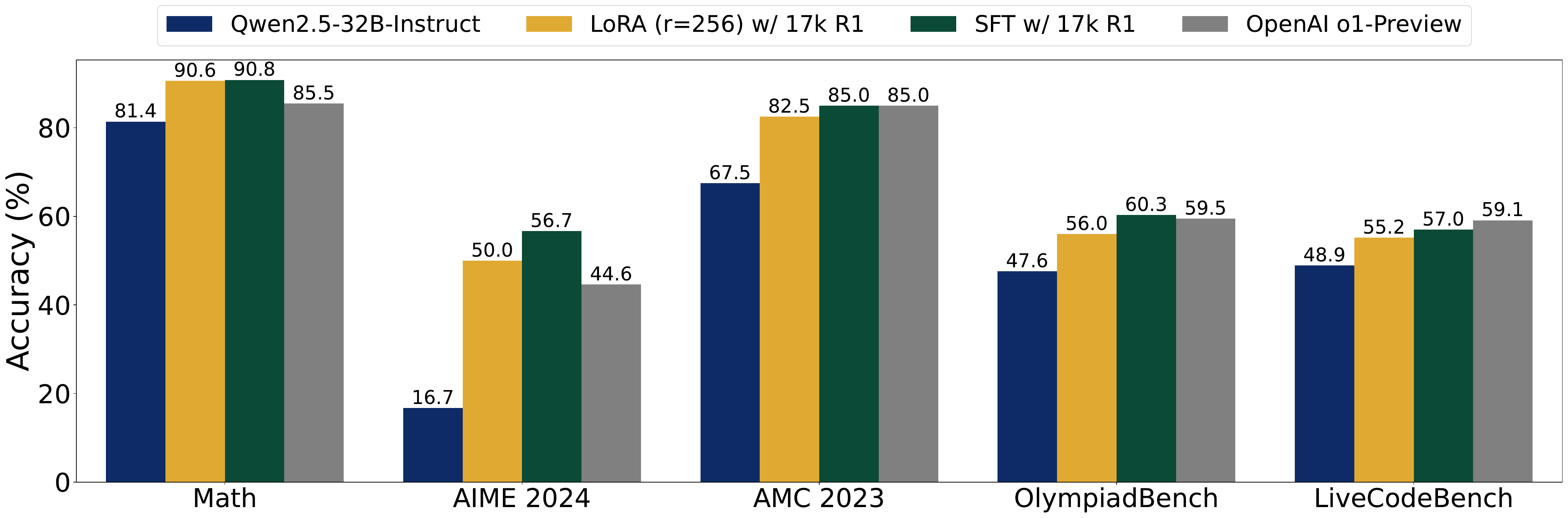}
        \caption{Performance of different models on five difference reasoning benchmarks.}
        \label{fig:intro-performance}
    \end{subfigure}
    \caption{\textbf{Learning to reason is data- and parameter-efficient.} When fine-tuned on a small amount (17k) of Long CoT samples distilled and reject-sampled from DeepSeek-R1 with either LoRA or full-parameter tuning, the model easily learns to perform reflection and backtracking by using keywords such as ``However'' and ``Alternatively'' (Top). Consequently, the fine-tuned models improve significantly across five popular math and coding benchmarks (Bottom). For fine-tuning, the base model is Qwen2.5-32B-Instruct.}
    \label{fig:longcot_teaser}
    \vspace{-2mm}
\end{figure*}

Large reasoning models (LRMs) leverage long chain-of-thoughts (Long CoTs) with reflection, backtracking, and self-validation to tackle challenging reasoning tasks~\citep{jaech2024openai, guo2025deepseek, team2024qwq}. 
However, the process of eliciting Long CoTs from available LLMs remains unclear, as existing methods are either closed-sourced~\citep{jaech2024openai, team2024qwq} or expensive to replicate~\citep{guo2025deepseek}.
In this paper, we first show that, surprisingly, an LLM can be cheaply and easily taught to produce Long CoT responses, significantly improving its reasoning capabilities. 
In particular, we find that this learning process can be both \textbf{data-efficient} and \textbf{parameter-efficient}.
By performing fully supervised fine-tuning (SFT) with only \textbf{17k} samples generated by DeepSeek R1, the Qwen2.5-32B-Instruct model achieves performance competitive with OpenAI o1-preview across a wide range of math and coding tasks~\citep{team2024qwq, yang2024qwen2, jaech2024openai}. 
In particular, it achieves 90. 8\% in Math-500 (+6.0\%), 56.7\% in AIME 2024 (+40.0\%), 85.0\% in AMC 2023 (+17.5\%), 60.3\% in OlympiadBench (+12.7\%) and 57.0\% in LiveCodeBench (+8.1\%)
~\citep{jain2024livecodebench}.
Even further, the model \textbf{can achieve o1-preview performance by updating fewer than $5\%$ parameters with LoRA fine-tuning}~\citep{hu2021lora}.
We show that the model successfully learns to reflect and revise its intermediate thoughts (e.g., frequently using reasoning keywords such as ``Alternatively'' and ``Wait, but'') and adopts long, coherent CoTs to tackle challenging problems (\cref{fig:longcot_teaser}).

Moreover, we identify the Long CoT \textit{structure} as the key characteristic of distilled data for eliciting strong reasoning performance rather than the \textit{specific contents} of individual reasoning steps within the Long CoT.
To test this, we conduct two sets of controlled studies by altering either the content of individual reasoning steps or the overall logical structure. 
To alter content, we perturb samples by replacing numbers with random digits or deleting reasoning keywords. Surprisingly, we find that these perturbations have little impact on the model performance: even when 50\% of numbers in training samples are randomly changed, the model only observes 3.3\% lower accuracy on the most challenging math benchmark, AIME 2024, as compared to training with correct samples.
To alter the global reasoning structure, we separate responses into reasoning steps and randomly shuffle, insert, or delete these steps. We observe that the trained model is much more sensitive to structural perturbations that break logical coherency in the long CoT. For example, when $67\%$ of the training samples' reasoning steps are shuffled, accuracy drops by $13.3\%$ on AIME 2024 problems relative to training with correct samples. 





In summary, our key contributions are:
 \begin{enumerate} 
 \item We demonstrate that an LLM can learn Long CoT reasoning in a data-efficient and parameter-efficient manner (i.e., LoRA). With fewer than 17k samples, we fine-tune the Qwen2.5-32B-Instruct model to be competitive with o1-preview. 
 \item We identify the structure of Long CoT as critical to the learning process rather than the content of individual reasoning steps. To validate this finding, we perform two groups of controlled experiments that modify either the structure or contents of samples.
 \item We conduct comprehensive ablations across model sizes and architectures, dataset sizes, data generation models (DeepSeek R1 and QwQ-32B-Preview), and on five popular math and coding reasoning benchmarks.
 \end{enumerate}

\section{Related work}
\textbf{Test Time Scaling for Large Language Models}
Scaling test-time compute has proven effective in enhancing the the reasoning capabilities of LLMs. 
This can be broadly categorized into two directions: single long CoT and repeatedly sampled CoT. The former trains models, such as OpenAI o1, DeepSeek R1, and Qwen QwQ, to generate individual, long CoT responses with in-context reflection and backtracking to handle complex reasoning tasks~\citep{guo2025deepseek, jaech2024openai, team2024qwq}. 
Alternatively, repeated sampling methods, such as Best-of-N or search-guided generation (e.g., MCTS), improve reasoning performance by invoking multiple responses from the model, sometimes guided by search algorithms and reward models~\citep{snell2024scaling, brown2024large}. In this paper, we focus on distilling the ability to generate individual, Long CoTs, and show it can be done in a data- and parameter-efficient manner. 

\textbf{Training to improve reasoning capabilities of LLMs}
LLM reasoning capabilities can be improved by approaches such as iterative self-improvement and reinforcement learning (RL)~\citep{zelikman2022star, lightman2023let, lambert2024t, yuan2024free, guo2025deepseek}. More recently, Tulu-3 introduces Reinforcement Learning with Verifiable Rewards (RLVR) to improve performance in tasks such as math and coding~\citep{hendrycksmath2021,jain2024livecodebench,numina_math_datasets}. PRIME proposes a RL-based method without process labels~\citep{yuan2024free}. The recent release of DeepSeek R1~\cite{guo2025deepseek} demonstrates that LLMs can learn to produce long CoT and improve reasoning using a pure RL-based approach. 
Instead of bootstrapping reasoning ability, this paper focuses on the surprising data- and parameter-efficiency of distilling reasoning abilities from an existing reasoning model to an LLM.

\paragraph{Distillation}
Distilling the outputs or logits generated by a larger or more capable model has become a standard technique to enhance model performance~\citep{hinton2015distilling}.  Typically, responses generated by higher-quality models are used to perform supervised fine-tuning on smaller models~\citep{lambert2024t}. 
The Vicuna model, for instance, demonstrates that ChatGPT-generated responses can be used to effectively and cheaply distill high-quality chatting capabilities~\citep{zheng2023judging}. In this paper, we show that \textbf{reasoning capabilities can also be cheaply distilled}. We note that concurrent work has also observed similar trends in distilling reasoning capability~\citep{min2024imitate, huang2024o1}. 
Our paper differs from these recent works by demonstrating that reasoning distillation can be achieved efficiently with minimal parameter updates. We also provide an in-depth analysis of the key factors driving reasoning improvements, including the roles of the reasoning structure and content, as well as comprehensive evaluations and ablations across different data sizes and teacher models.


\vspace{-1mm}
\section{Simple distillation is effective}
\label{sec:distill}
In this section, we present our distillation process and show that a small amount of \emph{well-curated} data, along with a simple parameter-efficient fine-tuning method (e.g., LoRA), can effectively improve reasoning capabilities in a large language model. 


\subsection{Experiments Setup}


\paragraph{Distillation data curation.} We use DeepSeek-R1~\citep{guo2025deepseek} and QwQ-32B-Preview~\citep{team2024qwq}, two open-source models with reasoning capabilities, to generate our distillation data. We select difficult prompts from the AMC/AIME~\footnote{\scriptsize These prompts are from previous years of competition, which do not include AIME 2024 and AMC 2023 in our evaluation suite.}, Math, and Olympiad subset from the Numina-Math dataset~\citep{numina_math_datasets}, as~\citet{min2024imitate} implies that hard problems can improve performance. We also incorporate coding problems from APPS~\citep{hendrycksapps2021} and TACO~\citep{li2023taco} datasets.
Specifically, we use GPT-4o-mini to classify the difficulty of the prompts according to the AoPS standard~\citep{achiam2023gpt}, and select math problems of difficulty higher than Leval 3, Olympiad higher than Level 8, and all AIME/AMC problems. We verify the correctness of the traces by checking against ground truth solutions using exact matching for math problems and code execution for coding problems. In total, we curated 12k math and 5k coding problems with correct responses from QwQ to serve as our training data. For R1 samples, we directly use the public R1-17k reasoning dataset\footnote{\scriptsize huggingface.co/datasets/bespokelabs/Bespoke-Stratos-17k.} that is curated following a similar procedure.

\textbf{Training details.} We perform training using Llama-Factory~\citep{zheng2024llamafactory}. We train the Qwen2.5-32B-Instruct using a batch size of 96, learning rate 1e-5 with a warm-up ratio of 0.1 and linear learning rate decay~\citep{yang2024qwen2}, following similar hyperparameters in~\citep{min2024imitate}. We use the next token prediction loss as the training objective~\citep{radford2018improving}. We use the same hyper-parameters except a 1e-4 learning rate for LoRA fine-tuning.

\textbf{Evaluation setup.}
We evaluate our models on five popular reasoning benchmarks for math and coding, including Math-500, OlympiadBench, AIME-2024\footnote{\scriptsize 
huggingface.co/datasets/AI-MO/aimo-validation-aime.}, AMC23\footnote{\scriptsize huggingface.co/datasets/AI-MO/aimo-validation-amc.}~\citep{hendrycksmath2021, he2024OlympiadBench} and LiveCodeBench~\citep{jain2024livecodebench}. For LiveCodeBench, we report a weighted average accuracy across its easy, medium, and hard difficulty levels.

\subsection{Key Insights}


\textbf{Small amount of data is enough.}
In \cref{fig:intro-performance}, we present the performance of models fine-tuned with the 17k R1 trained samples. Both the supervised fine-tuned (SFT) and LoRA fine-tuned models learn to generate Long CoT responses and improve significantly on all benchmarks with just 17k training samples. 
\begin{figure}[htbp]
\centering
\includegraphics[width=.45\textwidth]{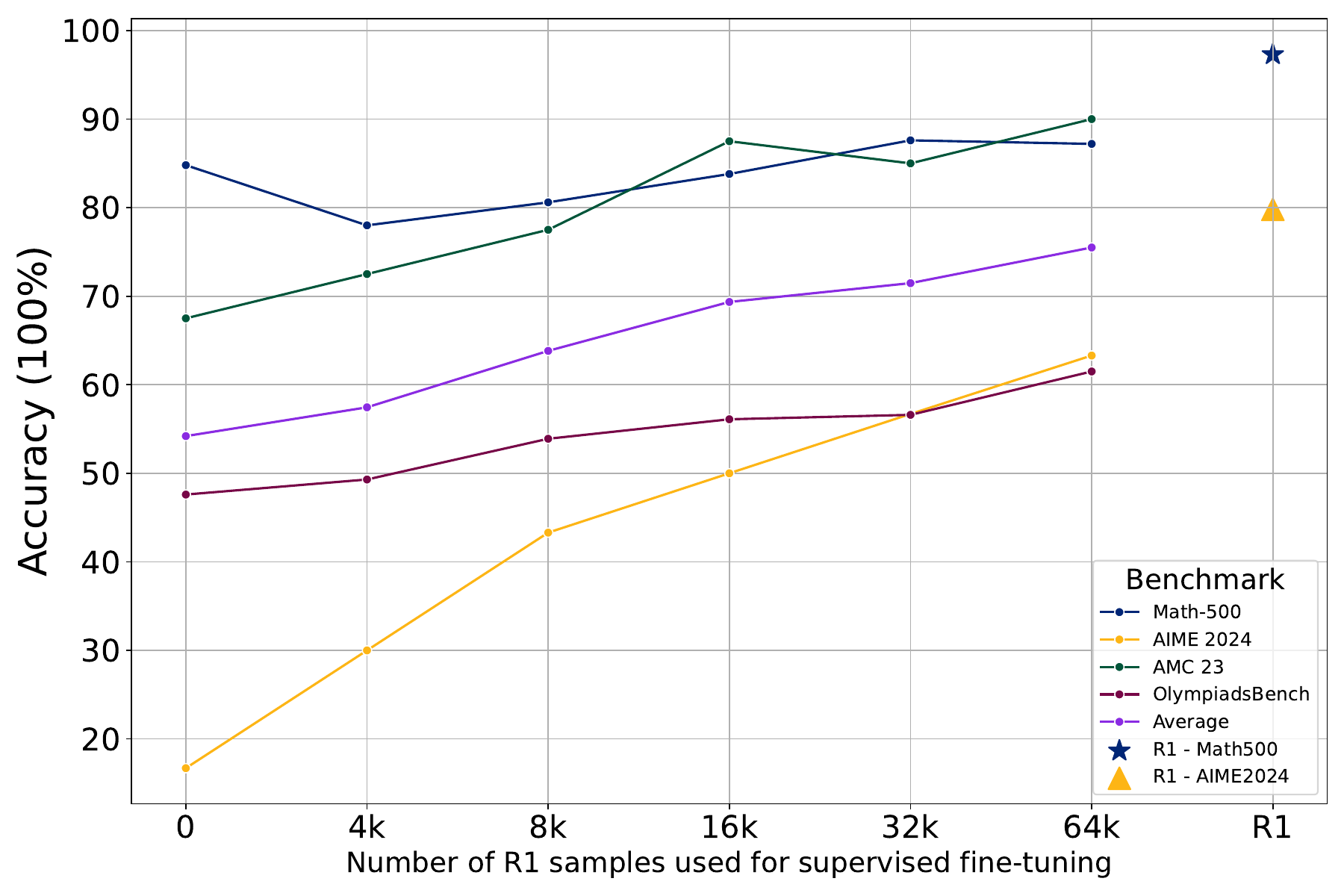}
\caption{\textbf{Model accuracy with different data sizes, and comparison to DeepSeek R1.} The teacher model is DeepSeek R1, and the student model is Qwen-32B-Instruct trained with full parameter fine-tuning. While the student model continues to benefits from more SFT data from DeepSeek R1, a small amount of data, e.g., 16k is sufficient to significantly boost the average performance by 15.2\%.}
\label{fig:data_scale}
\vspace{-4mm}
\end{figure}
We investigate the effect of distillation data size, ranging from 4k to 64k samples from R1, 
The results, presented in \cref{fig:data_scale}, shows that a small amount of data, e.g. 16k is enough to significantly improve the model performance (from average 54.2 to 69.4).


\paragraph{LoRA fine-tuning without performance degradation.} We next investigate the extent to which distilling Long CoT reasoning is knowledge-intensive. In addition to the results using 17k R1 samples as demonstrated in \cref{fig:intro-performance}, we also report the results for both SFT and LoRA fine-tuning with 7k and 17k QwQ samples in \cref{tab:sft-lora-qwq}.

Prior work~\citep{ghosh2024limitations, biderman2024lora} suggests that LoRA fine-tuning substantially under-performs full fine-tuning for knowledge-intensive tasks, and is limited to learning response initiation and style tokens. However, our results in \cref{fig:intro-performance} and \cref{tab:sft-lora-qwq} show that LoRA fine-tuned models achieve similar or even superior reasoning performance compared to full-parameter SFT across math and coding benchmarks. 
Additionally, we find that a model fine-tuned with LoRA using just 7k QwQ samples performs comparably to one trained on 17k QwQ-distilled samples. This demonstrates that reasoning distillation can be achieved efficiently with both minimal parameter updates and minimal data. 
As shown in \cref{fig:response_demo}, the LoRA fine-tuned model easily learns to generate Long CoT responses with reflection and self-verification. These observations suggest that Long CoT reasoning ability may not rely on deep knowledge acquisition but rather on learning structured reasoning patterns, which can be effectively distilled in a parameter-efficient manner. This also aligns with prior findings that methods such as Chain-of-Thought prompting elicit Short CoT reasoning primarily by shaping response structure rather than instilling deep factual knowledge~\citep{wei2022cot, yao2023tree_of_thoughts}. 

\begin{table}[ht!]
    \setlength{\tabcolsep}{4pt}
    \centering
    \caption{\textbf{Model accuracy with SFT and LoRA (rank=64).} Fine-tuning performed on Qwen2.5-32B-Instruct with QwQ samples. ``Olympiad.'' is short for ``OlympiadBench'', ``LCB.'' is short for ``LiveCodeBench''. We find that the learning process of Long CoT can be parameter efficient.}
    \label{tab:sft-lora-qwq}
    \resizebox{\linewidth}{!}{
    \begin{tabular}{lccccc}
        \toprule
        & \textbf{MATH500} & \textbf{AIME24} & \textbf{AMC23} & \textbf{Olympiad.} & \textbf{LCB.} \\
        \midrule
        Qwen2.5-32B-Inst. & 84.8 & 16.7 & 67.5 & 47.6 & 48.9 \\
        QwQ & 90.4 & 33.3 & 75.0 & 58.1 & 59.1 \\ 
        o1-preview & 85.5 & 44.6 & 87.5 & 59.2 & 59.1 \\
        \midrule
        \rowcolor{black!10} \multicolumn{6}{l}{\textbf{7k QwQ Samples}} \\
        SFT & 87.8 & 33.3 & 77.5 & 57.3 & 57.5 \\
        LoRA (r=64)   & 86.6 & 40.0 & 77.5 & 57.2 & 56.6 \\
        \midrule
        \rowcolor{black!10} \multicolumn{6}{l}{\textbf{17k QwQ Samples}} \\
        SFT & 87.8 & 33.3 & 70.0 & 56.7 & 57.9 \\
        LoRA (r=64)   & 86.6 & 33.3 & 90.0 & 56.0 & 56.2 \\
        
        \bottomrule
        
    \end{tabular}
    }
\end{table}





\section{Long CoT: Structure Is The Key}
\label{sec:key}
Motivated by the observation that fine-tuning with a small number of samples can significantly enhance model reasoning performance, we investigate the key factors driving this improvement. Specifically, we explore the contributions of two dimensions to the learning process:
\begin{enumerate}
    \item \textbf{The local content within a reasoning step}, including the correctness of the final answer, numbers in math derivations, and the use of reasoning keywords.
    \item \textbf{The global reasoning structure}, including reflection, self-validation, and backtracking across multiple reasoning steps to form a logically coherent long CoT.
\end{enumerate}

To understand their impact, we conduct two studies: (1) we perturb the content within individual reasoning steps -- such as the final answer, numerical digits, and reasoning keywords(\cref{sec:traces_change_contents}), and (2) we modify the global reasoning structure by inserting, deleting, and shuffling reasoning steps(\cref{sec:traces_change_structure}). We compare the performance of models trained on perturbed samples against both the base Qwen2.5-32B-Instruct model (i.e., Original) and model trained on correct, unperturbed samples (i.e., Correct), as shown in~\cref{tab:perturbations}. Our findings show that \textbf{the learning process is highly sensitive to modifications in the global reasoning structure, but remarkably tolerant to errors in the local contents.} 

\paragraph{Experiment setup} In this section, we use QwQ-32B-Preview to produce the distillation data and select a subset of 4618 correct responses as the training set (out of the 12k math data in~\cref{sec:distill}). All perturbations in this section are performed on this dataset unless otherwise stated. We train models on each separate variant of the dataset with the same hyperparameters as in~\cref{sec:distill} and report performance in~\cref{tab:perturbations}.

\subsection{Wrong or Corrupted Local Content}
\label{sec:traces_change_contents}
To study the importance of local content within individual steps, we preserve the overall reasoning structure while systematically perturbing the local content in training samples with different approaches. 





\paragraph{Wrong Answer Samples.} 
During our training data curation process in~\cref{sec:distill}, we only include samples that yield correct final answers. To assess whether correctness of the final answer is necessary for learning reasoning patterns, we instead train the model using an equivalent number of samples (4.6k) that lead to the \textit{wrong} answer. 
Surprisingly, we find that training the base model \textbf{without any samples that reach a correct final answer} still achieves an average accuracy of 63.1\% across benchmarks, only 3.2\% lower than training with entirely correct samples. 

\paragraph{Digits Corrupted Samples.} 
Building on the previous experiment, we next examine the role of correctness in the intermediate reasoning steps.
To evaluate this, we corrupt correct samples by replacing each digit with a random number between 0 and 9. Note that this is a severe corruption that can lead to nonsensical statements such as ``1+1=3''. Surprisingly, even when 70\% of the digits are corrupted, the model still maintains an average performance of 62\%, only 4.3\% below the correct sample baseline, demonstrating robustness to incorrect content. However, when all digits are corrupted, the average performance plunges to 2.7\%. 


\textbf{Reasoning Keyword Removal.}
Given the prevalence of reasoning \textit{keywords} in responses from LRMs (e.g., 'wait', 'let me think again', 'but'), one theory is that these specific phrases may invoke the reflection and back-tracking necessary to elicit strong reasoning performance. 
To evaluate it, we use GPT-4o-mini to identify sentences with occurrences of these reasoning keywords and randomly remove a fraction of them (e.g., 20\%, 50\%, 100\%). Our results show that even after removing all (100\%)    
  such keywords, the model still achieves an average accuracy of 63\%, which is within 3.3\% of accuracy from the model trained with correct samples. This suggests that these particular keywords do not fundamentally impact the model reasoning performance. 

\textbf{Conclusion.} We find that errors in local content -- such as incorrect mathematical derivations or missing reasoning keywords -- have minimal impact on overall performance. 

\begin{table}[t]
    \centering
    \caption{\textbf{Effect of trace perturbations on reasoning performance~\cref{sec:key}.} 
    All models are trained with base Qwen2.5-32B-Instruct. ``Olympiad.'' is short for ``OlympiadBench''. In particular, we study (1) traces with modified reasoning step contents: wrong answers, corrupted digits, and removed reasoning keywords, and (2) traces with modified structure: deleted, inserted, or shuffled steps. 
    \textbf{We find that structural perturbations are far more detrimental to model accuracy than content perturbations.}}
    \label{tab:perturbations}
    \footnotesize 
    \setlength{\tabcolsep}{3pt} 
    \resizebox{0.99\linewidth}{!}{
    \begin{tabular}{lccccc}
        \toprule
        & \textbf{MATH500} & \textbf{AIME24} & \textbf{AMC23} & \textbf{Olympiad.} & \textbf{Avg.} \\
        \midrule
        \rowcolor{black!10} \multicolumn{6}{l}{\textbf{Baselines}} \\
        Original & 84.8 & 16.7 & 67.5 & 47.6 & 56.7 \\
        Correct & 89.2 & 40.0 & 77.5 & 58.5 & 66.3 \\
        \midrule
        \rowcolor{black!10} \multicolumn{6}{l}{\textbf{Content Modifications}} \\
        \midrule
        \textbf{Wrong Answers} & 88.6 & 30.0 & 77.5 & 56.1 & 63.1 \\
        \midrule
        \textbf{Corrupted Digits} \\
        100\%  & 5.4  & 0.0  & 2.5  & 2.8  & 2.7 \\
        70\%   & 85.6 & 30.0 & 77.5 & 54.8 & 62.0 \\
        50\%   & 87.6 & 36.7 & 77.5 & 55.0 & 64.2 \\
        20\%   & 88.4 & 30.0 & 82.5 & 57.2 & 64.5 \\
        \midrule
        \textbf{Removed keywords} \\
        100\%    & 86.6 & 33.3 & 77.5 & 54.4 & 63.0 \\
        50\%     & 87.6 & 36.7 & 82.5 & 56.7 & 65.9 \\
        20\%     & 87.2 & 33.3 & 72.5 & 56.1 & 62.3 \\
        \midrule
        \rowcolor{black!10} \multicolumn{6}{l}{\textbf{Structure Modifications}} \\
        \midrule
        \textbf{Shuffled Steps} \\
        100\%    & 81.8 & 23.3 & 70.0 & 49.1 & 56.1 \\
        67\%     & 82.0 & 26.7 & 72.5 & 47.6 & 57.2 \\
        33\%     & 85.6 & 33.3 & 75.0 & 55.3 & 62.3 \\
        \midrule
        \textbf{Deleted Steps} \\
        100\%    & 79.2 & 13.3 & 60.0 & 45.4 & 49.5 \\
        67\%     & 84.2 & 26.7 & 55.0 & 48.1 & 53.5 \\
        33\%     & 88.2 & 23.3 & 80.0 & 57.7 & 62.3 \\
        \midrule
        \textbf{Inserted Steps} \\
        100\%    & 77.0 & 10.0 & 50.0 & 41.1 & 44.5 \\
        67\%     & 81.8 & 20.0 & 60.0 & 46.0 & 52.0 \\
        33\%     & 86.6 & 33.3 & 77.5 & 57.2 & 63.7 \\
        \bottomrule
    \end{tabular}
    }
    \vspace{-5mm}
\end{table}

\begin{figure*}[htbp]
\centering
\includegraphics[width=.85\textwidth]{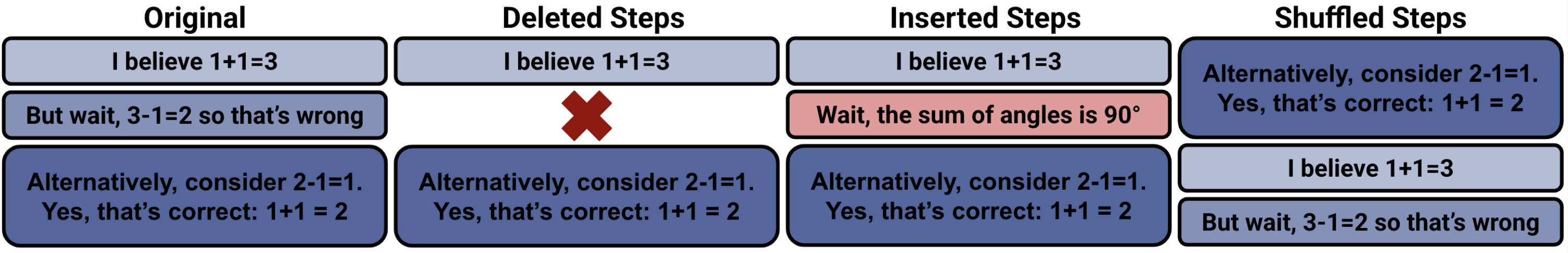}
\caption{\textbf{Reasoning step modifications.} To evaluate perturbations to global structure across reasoning steps, we perform three modifications: deletion, insertion, and shuffling. These modifications break logical consistency across steps and degrade model accuracy far more than changes to local content within reasoning steps.}
\label{fig:step_modifications}
\vspace{-4mm}
\end{figure*}

\subsection{Corrupted Global Reasoning Structure}
\label{sec:traces_change_structure}
Next, we examine the importance of reasoning \textit{structure} by performing three modifications to the reasoning traces: deletion, insertion, and shuffle. We first note that our system prompt (\cref{sec:system_prompt}) instructs the model to generate responses with thoughts enclosed in the tags `begin\_of\_thought' and `end\_of\_thought' and the final solution and step-by-step explanation in `begin\_of\_solution' and `end\_of\_solution'. All modifications are performed on the \textit{thoughts}, while the solution block is left unmodified.

We use Llama-3.3-70B-Instruct~\citep{dubey2024llama} to separate each reasoning trace into distinct reasoning steps, with boundaries determined by occurrences of backtracking, self-validation, reflection, or other breaks from a linear sequence of thoughts. We then generated nine modified variants of the dataset by applying each modification (insertion, deletion, and shuffle -- illustrated in~\cref{fig:step_modifications}) to 33\%, 67\%, or 100\% of reasoning steps in the 4,618 correct traces. Each variant is used to train the base model, Qwen2.5-32B-Instruct, and we report the resulting performance in \cref{tab:perturbations} and response lengths and reasoning keyword counts in~\cref{sec:lens_and_keyword_counts}.

\paragraph{Deleted reasoning steps.} As reasoning steps are increasingly deleted from the training data, model accuracy steadily declines and eventually regresses to the base model performance. Notably, retaining only the final solution and extensive step-by-step explanation (i.e., 100\% deletion case) does not suffice to learn strong reasoning capabilities. This suggests that correct long CoT demonstrations alone are insufficient. Instead, examples of handling errors and dead ends with backtracking, reflection, and self-validation are important for eliciting robust reasoning.

At 67\% deletion, the model imitates reasoning keywords (relative to the base model, keyword usage increases 45$\times$, and output token increases 9$\times$), but its accuracy does not improve accordingly. Consistent with ~\cref{sec:traces_change_contents}, this validates that merely adopting reasoning keywords and long responses is insufficient. We note, however, that as more steps are deleted, the response lengths also decrease significantly, which could contribute to reduced accuracy. We hypothesize that it is the breaking of logical consistency \textit{between} steps that causes accuracy degradation and validate this further in the following analysis. 

\paragraph{Inserted reasoning steps.} To further validate the importance of logical structure, we replace a subset of each trace's reasoning steps with a random sample of reasoning steps from other samples in the training set that lead to correct results. Unlike deletion, this approach generally preserves the original length of the reasoning trace, ensuring that accuracy degradation is not due simply to producing fewer steps. Relative to model variants trained with deleted reasoning steps, variants trained on inserted steps generate longer responses with more reasoning keywords, yet accuracy nonetheless deteriorates to, and even below, the level of the base model.

Interestingly, each inserted step is itself coherent and originates from a correct reasoning trace in the training data. Yet these internally-coherent steps appear in sequences that lack logical consistency and often from a separate domain (e.g., a combinatorics step may be inserted into a geometry solution), leading to contradictions and disjointed reflections. For instance, the model trained with inserted reasoning steps frequently references earlier steps that do not exist (e.g., {“Alternatively, consider a different approach” without specifying the prior approach) or enumerates edge cases in an inconsistent order (e.g., declaring a “Case 2” without “Case 1”). 

While the model readily produces coherent individual steps that reflect on a problem, the CoT fails to exhibit continuity \textit{across} reasoning steps. 
This aligns with the observations in the deletion setting: a mere increase in reasoning steps or keywords is insufficient for robust reasoning—logical consistency across steps is a critical factor.

\paragraph{Shuffled reasoning steps.} We next examine whether preserving the domain of each reasoning step, eliminating potential cross-domain confusion, but randomizing their order likewise impacts the model’s ability to reason. 

As the amount of shuffling increases, response length and reasoning keyword usage remain high, and in fact exceed the model trained on correctly ordered traces, yet accuracy declines sharply. Similar to the insertion experiments, the model imitates the syntax of per-step reasoning but loses logical consistency across steps. For instance, we find that over 92\% of model responses begin with a backtracking or self-validation keyword (e.g., \emph{``Alternatively,'' or ``Wait''}), even though there is no preceding content to correct or reconsider. The model also references prior calculations or cases that were never actually introduced in any preceding step. Thus, while the shuffled traces still contain valid domain-specific reasoning steps, their rearrangement leads to incoherent overall solutions. In other words, domain alignment alone does not prevent logical breakdown.

\paragraph{Conclusion.} Taken together, these findings show that providing error-free CoT demonstrations, increasing response lengths, imitating reasoning keywords and correct short CoT within individual steps, and preserving domain relevance for each step are \textit{not} sufficient to produce effective reasoning 
Further, our experiments on incorrect traces (\cref{sec:traces_change_contents}) demonstrate that learning reasoning capability is largely robust to local inaccuracies or miscalculations. Instead, global structural consistency is essential to elicit coherent long CoTs with the reflection, revision, and validation behaviors that produce strong reasoning performance.

\section{Ablation Study}
In this section, we conduct a series of ablation studies to answer the following questions:
\begin{enumerate}
    \item (\cref{sec:ablation_non_reasoning}) Does fine-tuning on Long CoT data lead to degraded performance on non-reasoning tasks?
    \item (\cref{sec:ablation_student}) How much does the Long CoT fine-tuning enhance the performance of different student models?
    \item (\cref{sec:ablation_test_time}) How does Long CoT model performance compare to the Best-of-N sampling performance of the base model?
    \item (\cref{sec:ablation_short}) How does Long CoT fine-tuning compare to Short CoT fine-tuning with the same dataset?
\end{enumerate}

\subsection{Performance on Non-Reasoning Benchmarks}
\label{sec:ablation_non_reasoning}

\begin{table}[ht!]
    \centering
    \caption{\textbf{Distilled Model Performance on Non-Reasoning Tasks.} The teacher model is  QwQ-32B-Preview, and the student model is Qwen2.5-32B-Instruct. Compared to QwQ, distilled models retain most of the base model's capabilities.}
    \label{tab:lora-non-reasoning}
    \footnotesize 
    \setlength{\tabcolsep}{6pt} 
    \resizebox{0.46\textwidth}{!}{
    \begin{tabular}{lcccc}
        \toprule
        & \textbf{MMLU} & \textbf{ARC-C} & \textbf{IEval}  & \textbf{MGSM}  \\
        \midrule
        Qwen2.5-32B-Inst. & 74.1 & 49.4 & 78.7 & 42.3 \\
        QwQ & 71.2 & 49.7  & 42.5 & 19.1 \\ 
        \midrule
        \rowcolor{black!10} \multicolumn{5}{l}{ \textbf{17k R1 Samples}} \\
        SFT & 73.0 & 49.0 & 77.8 & 33.7 \\
        LoRA (r=256)   & 75.5  & 47.3 & 78.4 & 38.7  \\
        \midrule
        \rowcolor{black!10} \multicolumn{5}{l}{ \textbf{17k QwQ Samples}} \\
        SFT & 78.4 & 49.5 & 75.8 & 33.0  \\
        LoRA (r=64)   &  78.5 & 46.7 & 74.1 & 30.6\\
        \midrule
        \rowcolor{black!10} \multicolumn{5}{l}{ \textbf{7k QwQ Samples}} \\
        SFT & 79.8 & 48.6 & 70.6 & 30.1 \\
        LoRA (r=64) & 79.1 & 47.4 & 75.4 & 31.1\\
        \bottomrule
    \end{tabular}
    }
\end{table}

While simple distillation enhances reasoning capabilities, it is essential to ensure that these improvements do not come at the cost of catastrophic forgetting or a decline in general language understanding and instruction-following abilities, which are crucial for broader task generalization. 

To assess this, we evaluate the performance of our SFT and LoRA fine-tuned models mentioned in~\cref{sec:distill} on a diverse set of benchmarks: MMLU (multi-task language understanding), ARC-C (science exam question), IEval (instruction-following), and MGSM (multilingual grade-school math problems)~\citep{hendrycks2021measuring,clark2018think,mitchell2023ieval,cobbe2021training}. 

As shown in~\cref{tab:lora-non-reasoning}, the base instruction model (Qwen2.5-32B-Instruct) performs well in all these tasks. The QwQ model, despite its strong reasoning capabilities, suffers significant degradation in 
instruction-following (i.e., 42.5\% on IEval) and multilingual tasks (i.e., 19.1\% on MGSM). 
In contrast, fine-tuning (through both SFT and LoRA) only on a small amount of Long CoT reasoning data from R1 or QwQ allows the distilled models to retain most of the base instruction model’s capabilities, avoiding the drastic performance drop seen in QwQ.

\subsection{Effect on Different Student Models}
\label{sec:ablation_student}

\begin{figure}[ht!]
    \centering
    \includegraphics[width=0.45\textwidth]{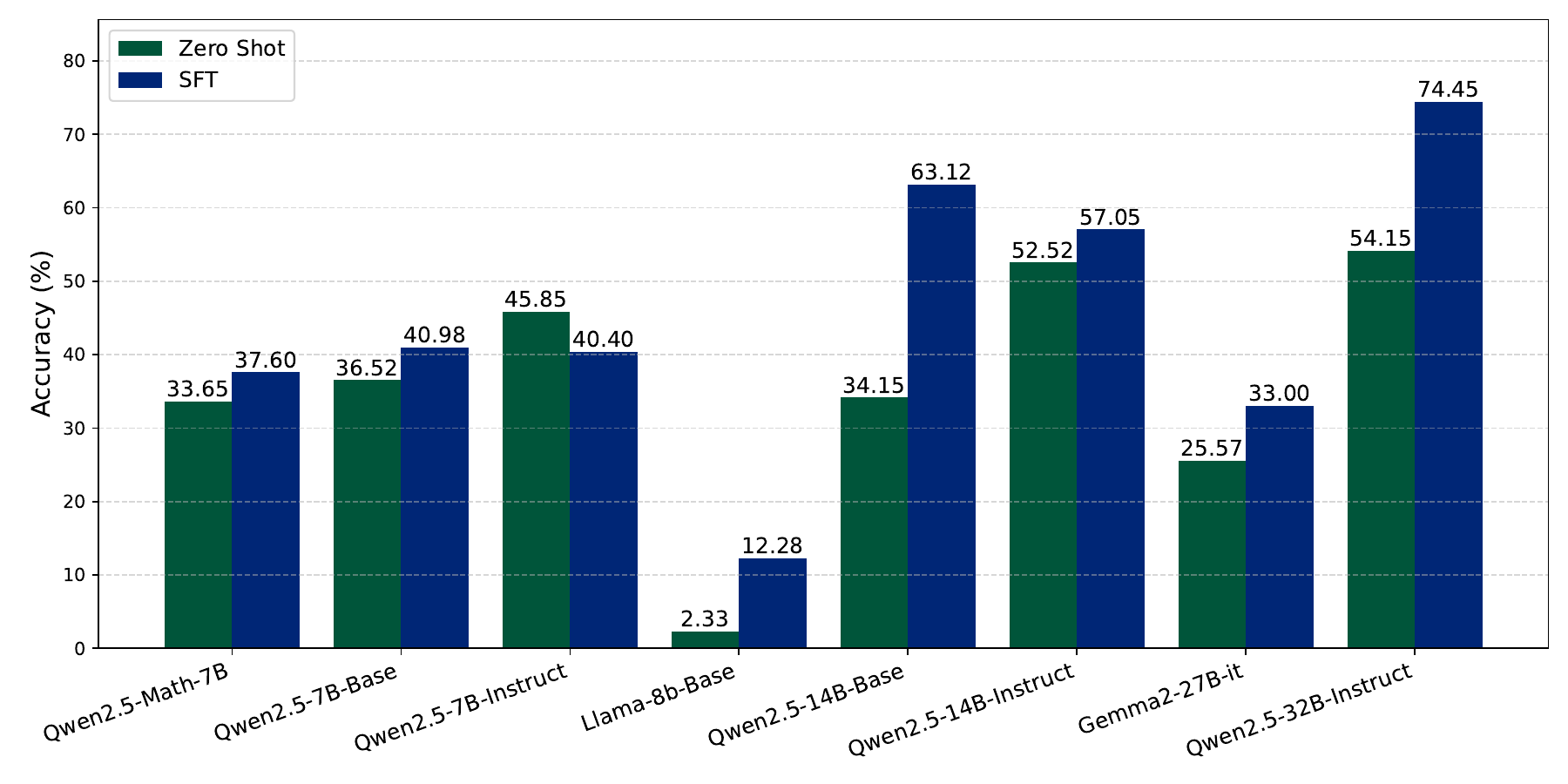}
    \caption{
    \textbf{Generalization to other models.} Accuracy for models of different sizes and architectures without SFT (green) and with SFT (blue). Most models show significant improvements when fine-tuned with 17k samples from R1-Preview, showing that the Long CoT fine-tuning is beneficial across models.} 
    \label{fig:other_models}
\end{figure}

In this section, we examine whether Long CoT reasoning capabilities can be elicited with different student models via fine-tuning (as described in~\cref{sec:distill}). 
Specifically, we train with the 17k samples on Qwen2.5-7B-Math, Qwen2.5-7-Base, Qwen2.5-7B-Instruct, Llama-3.1-8B, Qwen2.5-14B-Base, Qwen2.5-14B-Instruct, Gemma2-27B-it and Qwen2.5-32B-Instruct~\citep{yang2024qwen2, dubey2024llama, team2024gemma}. 
We find that seven out of eight models improve noticeably across multiple benchmarks, showing the effect of Long CoT as a general improvement across models.
However, not all models have showed the same degree of improvements as for Qwen2.5-32B-Instuct.
These findings suggest promising future directions for understanding the performance upper bound and data efficiency with various teacher and student models in the space of reasoning.




\subsection{Comparison to Best-of-N} 
\label{sec:ablation_test_time}
\begin{figure}[ht!]
    \centering
    \includegraphics[width=0.45\textwidth]{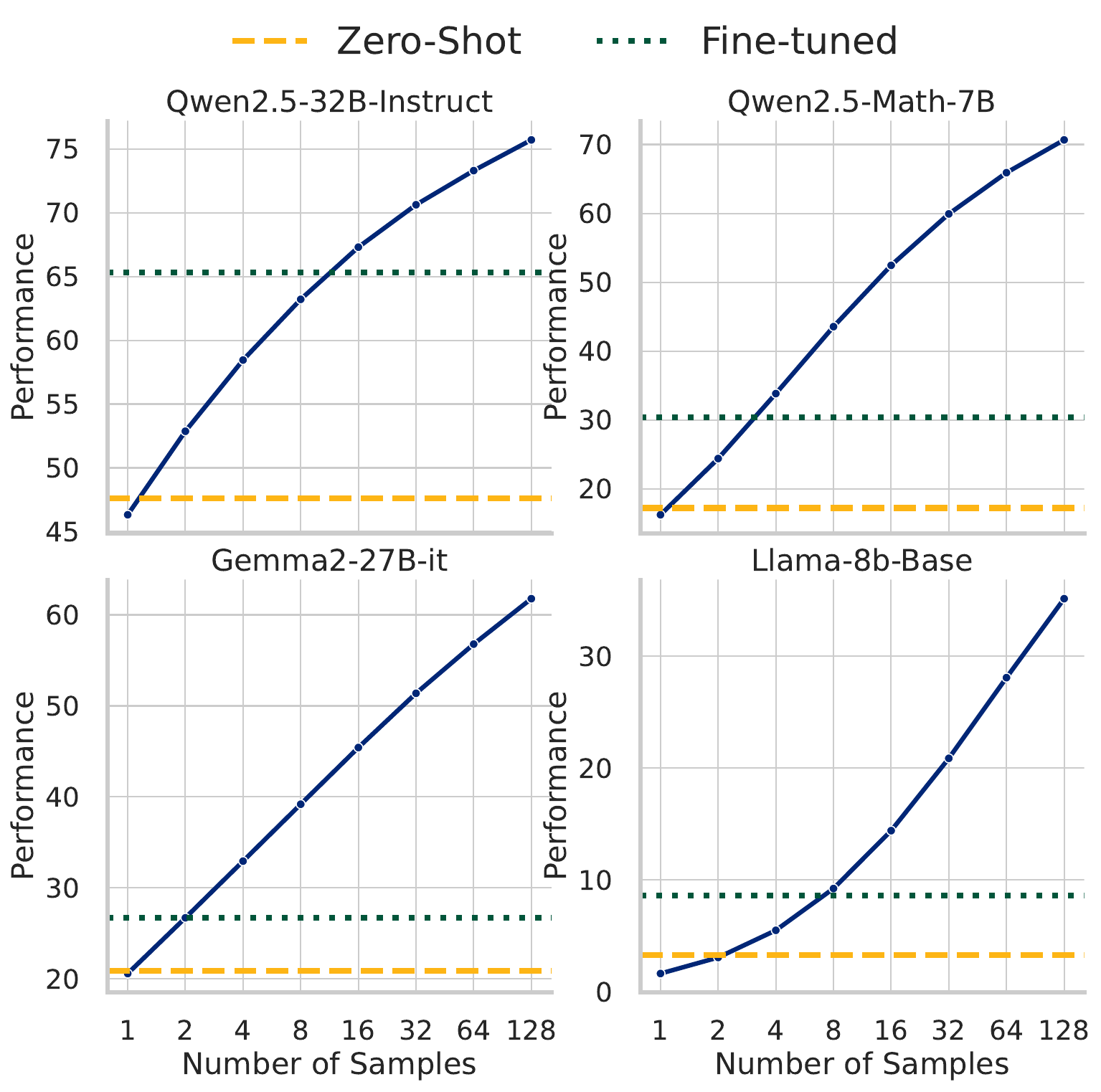}
    \caption{\textbf{SFT with Long CoT vs Best-of-N.} Accuracy of Qwen2.5-32B-Instruct before SFT (Zero-Shot), after SFT on 17k R1 samples (Fine-tuned), and Best-of-N samples on OlympiadBench. We find that fine-tuning on Long CoT achieves performance similar to Best of 2 to 16 samples.} 
    \label{fig:test_time}
\end{figure}
As discussed in~\cref{sec:ablation_student}, not all student models achieve significant performance improvements through Long CoT fine-tuning. We hypothesize that this variation is influenced by several factors, such as the extent to which the training data distribution differs from that of the student models and the inherent capabilities of the student models in these tasks. In this section, we compare the test-time scaling~\citep{ahn2024large, snell2024scaling} performance of the base model with its performance after Long CoT fine-tuning to understand the relationship between a model's ability to benefit from Long CoT fine-tuning and its intrinsic capabilities.

Specifically, we compare the performance of Long CoT fine-tuning against a Best-of-N sampling approach, where we generate 128 samples per prompt using an oracle verifier to select the best response. To introduce diversity, we employ a temperature of 0.5 and top-p sampling with a threshold of 0.8. The results, presented in~\cref{fig:test_time}, show that the Long CoT fine-tuned model performs comparably to Best-of-N sampling with 2 to 16 instances across all student models. 
Notably, the test-time scaling trends closely align with the improvements observed from Long CoT fine-tuning. For example, with eight parallel samples, Llama-3.1-8B achieves less than 10\% accuracy on OlympiadBench, and similarly, fine-tuning with correct Long CoT traces results in only marginal improvement. A comparable trend is observed in Gemma2-27B-it and Qwen2.5-Math-7B, reinforcing the relationship between test-time sampling efficiency and the benefits of Long CoT fine-tuning.

The performance of Best-of-N sampling continues to improve beyond 128 samples, suggesting that further gains are possible. This highlights the potential for enhancing Long CoT models through context scaling or by leveraging a broader range of reasoning paths inherent to the original model, potentially unlocking even higher performance.





\begin{table}[ht!]
\setlength{\tabcolsep}{6pt} 
\small\centering
\caption{
\textbf{Comparison of number of output tokens reasoning keywords, and the performance between training with Short or Long CoT.}
The original model is Qwen2.5-32B-Instruct. Benchmarks are ordered from easy to hard, where the model trained with Long CoT learns to produce longer CoTs and uses more keywords for harder problems.} 
\vspace{-1em}

\resizebox{0.46\textwidth}{!}{
\begin{tabular}{lccc}
\toprule
\textbf{Dataset} & \textbf{Original} & \textbf{Short CoT} & \textbf{Long CoT} \\
\midrule
\rowcolor{black!10} \multicolumn{4}{l}{\textit{Avg. output tokens}} \\
MATH500           & 684   & 515   & 3972  \\
AMC23       & 728   & 605   & 5037  \\
OlympiadBench & 1275  & 948   & 8616  \\
AIME24      & 825   & 687   & 15902  \\
\midrule
\rowcolor{black!10} \multicolumn{4}{l}{\textit{Avg. keywords per response}} \\
MATH500          & 0.00  & 0.00  & 41.75  \\
AMC23      & 0.00  & 0.00  & 39.20  \\
OlympiadBench & 0.01  & 0.01  & 97.20  \\
AIME24      & 0.00  & 0.07  & 260.90  \\
\rowcolor{black!10} \multicolumn{4}{l}{\textit{Performance}} \\
MATH500          & 84.8  & 70.4 (-14.4) & \textbf{89.2} (+4.4) \\
AMC23      & 67.5  & 55.0 (-12.5) & \textbf{77.5} (+10.0) \\
OlympiadBench & 47.6  & 36.4 (-11.2) & \textbf{58.5} (+10.9) \\
AIME24      & 16.7  & 13.3 (-3.4) & \textbf{40.0} (+23.3) \\
\bottomrule
\end{tabular}
}
\label{tab:longcot_stats}
\end{table}
\subsection{Comparison to Short CoT Fine-tuning}
\label{sec:ablation_short}
In this section, we provide a direct comparison to training with short CoT. In particular, we compare results training on the 4.6k samples of Long CoT generated by QwQ-32B-Preview (\cref{sec:key}), and the short CoT denoted in the NuminaMath-CoT dataset~\citep{numina_math_datasets}. \cref{tab:longcot_stats} summaries the statistics. Training with Long CoT enables the model to use more reasoning keywords (full list in~\cref{sec:keywords}), produces longer responses to harder problems, and is the key to improved performance.

\section{Conclusion}
Large reasoning models unlock new capabilities by using a longer chain of thoughts that involves reflection and backtracking to answer challenging problems. In this paper, we show that such capability can be easily fine-tuned using a few thousand examples and with low-rank adapters. We further show that the key to the learning process is the logical structure of the samples rather than the content of individual reasoning steps. Finally, we discuss several ablations with various teacher-student models and compare them to the best-of-N approach. Together, our work deepens the understanding of what is needed to instill large language models with strong reasoning capabilities and identify potential future directions.


\section*{Impact Statement}
This paper aims to contribute to the advancement of Machine Learning. While our work may have various societal implications, we do not find any that require specific emphasis currently.

\section*{Acknowledgement}
This work is funded by the Sky Computing Lab at UC Berkeley. We extend our sincere gratitude to Sumanth Hegde, Eric Tang, and Kourosh Hakhamaneshi from AnyScale for their invaluable feedback, assistance in building the evaluation infrastructure, and and contributions to the best-of-n experiments. We are grateful for the generous compute resources support from Databricks, Lambda Labs, and Anyscale. In particular, we thank Jonathan Frankle (Databricks) and Chuan Li (Lambda Labs) for facilitating access to these resources.

\nocite{langley00}

\bibliography{example_paper}
\bibliographystyle{icml2025}

\newpage
\appendix
\onecolumn
\section{Short and Long CoT Response Examples}
\begin{tcolorbox}[colback=white,colframe=gray,title=Question]
For how many integers \( n \) does the expression
\[
\sqrt{\frac{\log (n^2) - (\log n)^2}{\log n - 3}}
\]
represent a real number, where \(\log\) denotes the base \(10\) logarithm?
\end{tcolorbox}

\begin{tcolorbox}
[colback=white,colframe=gray,title=Short CoT answer (1/2)]
\label{ref:figure_shortcot}
To solve the problem, we start by analyzing the given expression and the conditions under which it represents a real number. The expression is:

\[
\sqrt{\frac{\log (n^2) - (\log n)^2}{\log n - 3}}
\]

For the square root to represent a real number, the argument of the square root, which is the fraction inside, must be nonnegative. Additionally, since we are dealing with logarithms, \( n \) must be a positive integer. Let's break down the solution into detailed steps:

1. Condition for the Denominator
The denominator of the fraction is \( \log n - 3 \). For the fraction to be nonnegative, we need \( \log n - 3 > 0 \). This implies that \( \log n > 3 \), which translates to 

\[
n > 10^3 = 1000.
\]

Therefore, \( n \) must be greater than \( 1000 \).

2. Simplifying the Numerator
The numerator of the fraction is \( \log (n^2) - (\log n)^2 \). Using logarithmic properties, we can rewrite \( \log (n^2) \) as \( 2\log n \). Thus, the numerator becomes 

\[
2\log n - (\log n)^2.
\]

3. Analyzing the Fraction
Substituting the simplified numerator into the fraction, we get:

\[
\frac{2\log n - (\log n)^2}{\log n - 3}
\]

For this fraction to be nonnegative, given that the denominator is positive (from step 1), the numerator \( 2\log n - (\log n)^2 \) must also be nonnegative. This leads to the inequality:

\[
2\log n - (\log n)^2 \geq 0.
\]

4. Solving the Inequality
The inequality 

\[
2\log n - (\log n)^2 \geq 0
\]

can be rewritten as 

\[
(\log n)^2 - 2\log n \leq 0.
\]

Factoring,

\[
\log n (\log n - 2) \leq 0.
\]

Solving this inequality, we find that 

\[
0 \leq \log n \leq 2.
\]

Converting back to \( n \), this implies:

\[
1 \leq n \leq 100.
\]

5. Combining Conditions
From step 1, we have \( n > 1000 \), and from step 4, we have \( 1 \leq n \leq 100 \). However, these two conditions are contradictory. The correct interpretation should be that \( n \) must satisfy both the condition of being greater than \( 1000 \) (from the denominator analysis) and the condition derived from the inequality, which was misinterpreted in the combination step. The correct combination should focus on the range derived from the inequality, which is \( 1 \leq n \leq 100 \), and disregard the incorrect application of the denominator condition in this context.
\end{tcolorbox}
\begin{tcolorbox}[colback=white,colframe=gray,title=Short CoT answer (2/2)]
6. Conclusion
The integers \( n \) that satisfy the given conditions are those for which 

\[
1 \leq n \leq 100.
\]

Counting these integers, we find there are 

\[
\boxed{100}
\]

such integers.
\end{tcolorbox}

\begin{tcolorbox}[colback=white,colframe=gray,title=Long CoT answer (1/4)]
\label{ref:figure_longcot}
\texttt{\textless|}begin\_of\_thought\texttt{|>}
So I've got this problem here:
\[
\sqrt{\frac{\log (n^2) - (\log n)^2}{\log n - 3}}
\]

And I need to find out for how many integer values of \( n \) this expression represents a real number. The logarithm here is base 10.

First off, since it's a square root, the expression inside the square root has to be non-negative for the whole thing to be a real number. So, that means:

\[
\frac{\log (n^2) - (\log n)^2}{\log n - 3} \geq 0
\]

Additionally, the argument of the logarithm has to be positive because logarithms are only defined for positive real numbers. So, \( n^2 > 0 \), which is true for all \( n \neq 0 \), and \( n > 0 \) since \( n^2 > 0 \) is always true for \( n \neq 0 \), but we need to ensure \( n > 0 \) because the logarithm of a negative number is undefined in real numbers.

So, \( n \) has to be a positive integer.

Now, let's simplify the expression inside the square root.

I know that \( \log(n^2) = 2 \log n \), because of the logarithm power rule. So, substituting that in:

\[
\frac{2 \log n - (\log n)^2}{\log n - 3}
\]

Let me denote \( \log n = x \) to make it easier. So, the expression becomes:

\[
\frac{2x - x^2}{x - 3}
\]

I can factor the numerator:

\[
2x - x^2 = x(2 - x)
\]

So, the expression is:

\[
\frac{x(2 - x)}{x - 3}
\]

I need to analyze when this fraction is non-negative.

To do that, I can look at the sign of each factor:

- \( x = \log n \)

- \( 2 - x = 2 - \log n \)

- \( x - 3 = \log n - 3 \)

So, the expression is:

\[
\frac{\log n (2 - \log n)}{\log n - 3}
\]

I need to determine when this fraction is greater than or equal to zero.

Let me consider the critical points where each factor is zero:

1. \( \log n = 0 \) \( \Rightarrow n = 10^0 = 1 \)

2. \( 2 - \log n = 0 \) \( \Rightarrow \log n = 2 \) \( \Rightarrow n = 10^2 = 100 \)

3. \( \log n - 3 = 0 \) \( \Rightarrow \log n = 3 \) \( \Rightarrow n = 10^3 = 1000 \)

These points divide the number line into intervals where the expression doesn't change sign.

So, the intervals to consider are:

- \( 1 < n < 100 \)

- \( 100 < n < 1000 \)

- \( n > 1000 \)
\end{tcolorbox}

\begin{tcolorbox}[colback=white,colframe=gray,title=Long CoT answer (2/4)]
Wait, but I need to consider \( n \) as positive integers, and \( n \geq 1 \).

But actually, \( n \) has to be at least 1, but \( n = 1 \) makes \( \log n = 0 \), which is a critical point.
So, let's consider the intervals:

- \( 1 < n < 100 \)

- \( n = 100 \)

- \( 100 < n < 1000 \)

- \( n = 1000 \)

- \( n > 1000 \)

I need to test the sign of the expression in each interval.

First, let's consider \( n = 1 \):

\( \log 1 = 0 \), so the numerator is \( 0 \), and the denominator is \( 0 - 3 = -3 \), so the fraction is \( 0 / -3 = 0 \), which is acceptable since it's zero.

Next, for \( 1 < n < 100 \):

Let's pick \( n = 10 \), for example.

\( \log 10 = 1 \)

So, \( x = 1 \)

Then, \( 2 - x = 1 \)

\( x - 3 = -2 \)

So, \( \frac{1 \cdot 1}{-2} = -\frac{1}{2} < 0 \), which is not acceptable.

Wait, but earlier I thought the expression inside the square root needs to be non-negative. So, this interval doesn't work.

Next interval: \( n = 100 \)

\( \log 100 = 2 \)

Numerator: \( 2 \cdot (2 - 2) = 0 \)

Denominator: \( 2 - 3 = -1 \)

So, \( 0 / -1 = 0 \), which is acceptable.

Then, \( 100 < n < 1000 \)

Let's pick \( n = 1000^{0.5} = 10^{2.5} \approx 316.2 \), but since \( n \) is integer, let's take \( n = 300 \)

\( \log 300 \approx 2.477 \)

So, \( x \approx 2.477 \)

\( 2 - x \approx -0.477 \)

\( x - 3 \approx -0.523 \)

So, \( \frac{2.477 \cdot (-0.477)}{-0.523} \approx \frac{-1.182}{-0.523} \approx 2.26 > 0 \), which is acceptable.

Next, \( n = 1000 \)

\( \log 1000 = 3 \)

Denominator: \( 3 - 3 = 0 \), so the expression is undefined. Not acceptable.

Finally, \( n > 1000 \)

Let's pick \( n = 10000 \)

\( \log 10000 = 4 \)

So, \( x = 4 \)

\( 2 - x = -2 \)

\( x - 3 = 1 \)

So, \( \frac{4 \cdot (-2)}{1} = -8 < 0 \), which is not acceptable.

So, summarizing:

- \( n = 1 \): acceptable

- \( 1 < n < 100 \): not acceptable

- \( n = 100 \): acceptable

- \( 100 < n < 1000 \): acceptable

- \( n = 1000 \): not acceptable

- \( n > 1000 \): not acceptable

Therefore, the acceptable ranges are:

- \( n = 1 \)

\end{tcolorbox}

\begin{tcolorbox}
[colback=white,colframe=gray,title=Long CoT answer (3/4)]

- \( n = 100 \)

- \( 100 < n < 1000 \)

Wait, but \( n = 100 \) is included in \( 100 < n < 1000 \)? No, \( n = 100 \) is a separate point.

Actually, \( n = 100 \) is the upper limit of \( 100 < n < 1000 \), but since \( n = 100 \) is acceptable, I need to include it.

Wait, no, \( n = 100 \) is a separate point where the expression is zero, which is acceptable.

So, the acceptable \( n \) are:

- \( n = 1 \)

- \( 100 \leq n < 1000 \)

Wait, but earlier I thought \( n = 1000 \) is not acceptable, but \( n = 1000 \) is not included in \( 100 \leq n < 1000 \), which is correct.

So, now, I need to count the integer values of \( n \) in these ranges.

First, \( n = 1 \): 1 value

Second, \( 100 \leq n < 1000 \): that's from 100 to 999, inclusive.

So, \( 999 - 100 + 1 = 900 \) values.

Therefore, total integers are \( 1 + 900 = 901 \).

But wait, earlier I thought \( n = 100 \) is acceptable, and it is included in \( 100 \leq n < 1000 \), so no need to count it separately.

Wait, no, \( n = 100 \) is included in \( 100 \leq n < 1000 \), which already includes 100 to 999, so total is 900 values.

Plus \( n = 1 \), so total is 901.

But, I need to make sure that for all \( n \) in \( 100 \leq n < 1000 \), the expression is positive.

Wait, except for \( n = 100 \), where it's zero, which is acceptable.

But in the interval \( 100 < n < 1000 \), it's positive, as I tested earlier.

So, yes, total is 901.

Therefore, the number of integer values of \( n \) that satisfy the condition is 901.

**Final Answer**

\[
\boxed{901}
\]

\texttt{\textless|}end\_of\_thought\texttt{|>}

\texttt{\textless|}begin\_of\_solution\texttt{|>}

To determine for how many integer values of \( n \) the expression

\[
\sqrt{\frac{\log (n^2) - (\log n)^2}{\log n - 3}}
\]

represents a real number, we first note that the expression inside the square root must be non-negative:

\[
\frac{\log (n^2) - (\log n)^2}{\log n - 3} \geq 0
\]

Additionally, since logarithms are only defined for positive real numbers, we require \( n > 0 \).

1. Simplifying the expression

Using the logarithm power rule, we have:

\[
\log(n^2) = 2 \log n
\]

Thus, the expression becomes:

\[
\frac{2 \log n - (\log n)^2}{\log n - 3}
\]

Let \( x = \log n \). Then, the expression simplifies to:

\[
\frac{x(2 - x)}{x - 3}
\]

\end{tcolorbox}

\begin{tcolorbox}
[colback=white,colframe=gray,title=Long CoT answer (4/4)]

2. Finding critical points

The critical points are where the numerator or denominator equals zero:

- \( x = 0 \) \( \Rightarrow n = 1 \)
- \( 2 - x = 0 \) \( \Rightarrow x = 2 \Rightarrow n = 100 \)
- \( x - 3 = 0 \) \( \Rightarrow x = 3 \Rightarrow n = 1000 \)

3. Analyzing intervals

The intervals to consider are:

- \( 1 < n < 100 \)
- \( n = 100 \)
- \( 100 < n < 1000 \)
- \( n = 1000 \)
- \( n > 1000 \)

Testing the sign of the expression in each interval:

- For \( n = 1 \): The expression equals 0, which is acceptable.
- For \( 1 < n < 100 \): The expression is negative, not acceptable.
- For \( n = 100 \): The expression equals 0, acceptable.
- For \( 100 < n < 1000 \): The expression is positive, acceptable.
- For \( n = 1000 \): The expression is undefined, not acceptable.
- For \( n > 1000 \): The expression is negative, not acceptable.

4. Counting acceptable integers

The acceptable ranges are:

- \( n = 1 \)
- \( 100 \leq n < 1000 \)

The number of integers in \( 100 \leq n < 1000 \) is \( 900 \) (from 100 to 999 inclusive). Adding \( n = 1 \) gives a total of:

\[
1 + 900 = 901
\]

Thus, the number of integer values of \( n \) that satisfy the condition is:

\[
\boxed{901}
\]
\texttt{\textless|}end\_of\_solution\texttt{|>}
\end{tcolorbox}

\section{Reasoning keywords}
\label{sec:keywords}
We create a bank of reasoning keywords by manual inspection of reasoning responses from QwQ and R1. We identify these following words and phrases as frequent indicators of reflection, backtracking, and self-validation: ``Alternatively'', ``Wait'', ``Just to be thorough'', ``Just to make sure'', ``Let me just double-check'', ``Let me try another'', ``Let me verify'', ``Let me check'', ``Hmm'', ``But'', ``Maybe I should consider'', ``Maybe I can consider''.

\section{Long CoT System Prompt}
\label{sec:system_prompt}
We adopt the following system prompt from Still-2~\citep{min2024imitate}:
\begin{tcolorbox}
[colback=white,colframe=gray,title=Long CoT System Prompt]
Your role as an assistant involves thoroughly exploring questions through a systematic long thinking process before providing the final precise and accurate solutions. This requires engaging in a comprehensive cycle of analysis, summarizing, exploration, reassessment, reflection, backtracking, and iteration to develop well-considered thinking process. Please structure your response into two main sections: Thought and Solution. In the Thought section, detail your reasoning process using the specified format: \texttt{\textless |}begin\_of\_thought\texttt{|\textgreater} thought with steps separated with \texttt{\textbackslash n\textbackslash n}\} \texttt{\textless |}end\_of\_thought\texttt{|\textgreater} Each step should include detailed considerations such as analyzing questions, summarizing relevant findings, brainstorming new ideas, verifying the accuracy of the current steps, refining any errors, and revisiting previous steps. In the Solution section, based on various attempts, explorations, and reflections from the Thought section, systematically present the final solution that you deem correct. The solution should remain a logical, accurate, concise expression style and detail necessary step needed to reach the conclusion, formatted as follows: \texttt{\textless |}begin\_of\_solution\texttt{|\textgreater} {final formatted, precise, and clear solution} \texttt{\textless |}end\_of\_solution\texttt{|\textgreater} Now, try to solve the following question through the above guidelines:
\end{tcolorbox}

\section{Average response lengths and keyword counts}
\label{sec:lens_and_keyword_counts}
\begin{table}[h]
\setlength{\tabcolsep}{6pt} 
\small\centering
\caption{Average keyword counts and output tokens for deleted steps.}
\vspace{-1em}
\begin{tabular}{lcccc}
\toprule
\textbf{Dataset} & \textbf{0\%} & \textbf{33\%} & \textbf{67\%} & \textbf{100\%} \\
\midrule
\rowcolor{black!10} \multicolumn{5}{l}{\textit{Avg. output tokens}} \\
Math           & 3551   & 2979   & 2078  & 482 \\
AMC 2023       & 4838  & 6612   & 4623  & 609 \\ 
OlympiadBench & 7234   & 6802   & 4978 & 595 \\ 
AIME 2024      & 13088   & 11889   & 6798  & 620 \\
\midrule
\rowcolor{black!10} \multicolumn{5}{l}{\textit{Avg. keywords per response}} \\
Math           & 32   & 28   & 20  & 0.017 \\
AMC 2023       & 39  & 85.6   & 77.8  & 0 \\ 
OlympiadBench & 77   & 70   & 56 & 0.009 \\ 
AIME 2024      & 143   & 143   & 90  & 0 \\
\bottomrule
\end{tabular}
\label{tab:delete_stats}
\end{table}

\begin{table}[h]
\setlength{\tabcolsep}{6pt} 
\small\centering
\caption{Average keyword counts and output tokens for inserted steps.}
\vspace{-1em}
\begin{tabular}{lcccc}
\toprule
\textbf{Dataset} & \textbf{0\%} & \textbf{33\%} & \textbf{67\%} & \textbf{100\%} \\
\midrule
\rowcolor{black!10} \multicolumn{5}{l}{\textit{Avg. output tokens}} \\
Math           & 3551   & 4189   & 3900  & 5383 \\
AMC 2023       & 4838  & 7089   & 5464  & 5137 \\ 
OlympiadBench & 7234   & 7558   & 6990 & 5407 \\ 
AIME 2024      & 13088   & 12858   & 12864  & 5304 \\
\midrule
\rowcolor{black!10} \multicolumn{5}{l}{\textit{Avg. keywords per response}} \\
Math           & 32   & 39   & 39  & 41 \\
AMC 2023       & 39  & 98   & 44  & 35 \\ 
OlympiadBench & 77   & 76   & 80 & 38 \\ 
AIME 2024      & 143   & 127   & 165  & 44 \\
\bottomrule
\end{tabular}
\label{tab:insert_stats}
\end{table}

\begin{table}[h]
\setlength{\tabcolsep}{6pt} 
\small\centering
\caption{Average keyword counts and output tokens for shuffled steps.}
\vspace{-1em}
\begin{tabular}{lcccc}
\toprule
\textbf{Dataset} & \textbf{0\%} & \textbf{33\%} & \textbf{67\%} & \textbf{100\%} \\
\midrule
\rowcolor{black!10} \multicolumn{5}{l}{\textit{Avg. output tokens}} \\
Math           & 3551   & 4284   & 5784  & 5613 \\
AMC 2023       & 4838  & 6802   & 10198  & 8661 \\ 
OlympiadBench & 7234   & 8942   & 12154 & 12167 \\ 
AIME 2024      & 13088   & 13451   & 16221  & 18054 \\
\midrule
\rowcolor{black!10} \multicolumn{5}{l}{\textit{Avg. keywords per response}} \\
Math           & 32   & 45   & 61  & 70 \\
AMC 2023       & 39  & 65   & 74  & 67 \\ 
OlympiadBench & 77   & 111   & 166 & 137 \\ 
AIME 2024      & 143   & 161   & 201  & 210 \\
\bottomrule
\end{tabular}
\label{tab:shuffle_stats}
\end{table}

\end{document}